\documentclass{article}

\usepackage{PRIMEarxiv}

\usepackage[utf8]{inputenc} 
\usepackage[T1]{fontenc}    
\usepackage{hyperref}       
\usepackage{url}            
\usepackage{booktabs}       
\usepackage{amsfonts}       
\usepackage{nicefrac}       
\usepackage{microtype}      
\usepackage{lipsum}
\usepackage{fancyhdr}       
\usepackage{graphicx}       
\graphicspath{{media/}}     

\usepackage{algorithm}
\usepackage{algorithmic}
\usepackage{amsmath,amssymb,amsfonts}
\usepackage{graphicx}
\usepackage{textcomp}
\usepackage{xcolor}
\usepackage{multicol}

\title{Towards efficient deep autoencoders for multivariate time series anomaly detection 
\thanks{\textit{\underline{Citation}}: 
\textbf{Authors. Title. Pages.... DOI:000000/11111.}} 
}

\author{
  Marcin Pietron \\
  Institute of Electronics \\
  AGH-UST \\
  Cracow\\
  \texttt{pietron@agh.edu.pl} \\
  \and
  Dominik Zurek \\
  Institute of Computer Science \\
  AGH-UST \\
  Cracow\\
  \texttt{pietron@agh.edu.pl} \\
  \and
  Kamil Faber \\
  Institute of Computer Science \\
  AGH-UST \\
  Cracow\\
  \texttt{pietron@agh.edu.pl} \\
  \texttt{} \\
  \and
  Roberto Corizzo \\
  Department of Computer Science \\
  American University \\
  Washington DC\\
  \texttt{rcorizzo@american.edu} \\
}

\begin{document}
\maketitle

\begin{abstract}
Multivariate time series anomaly detection is a crucial problem in many industrial and research applications. Timely detection of anomalies allows, for instance, to prevent defects in manufacturing processes and failures in cyberphysical systems.  
Deep learning methods are preferred among others for their accuracy and robustness for the analysis of  complex multivariate data. 
However, a key aspect is being able to extract predictions in a timely manner, to accommodate real-time requirements in different applications.
In the case of deep learning models, model reduction is extremely important to achieve optimal results in real-time systems with limited time and memory constraints.
In this paper, we address this issue by proposing a novel compression method for deep autoencoders that involves three key factors. First, pruning reduces the number of weights, while preventing catastrophic drops in accuracy by means of a fast search process that identifies high sparsity levels. Second, linear and non-linear quantization reduces model complexity by reducing the number of bits for every single weight. 
The combined contribution of these three aspects allow the model size to be reduced, by removing a subset of the weights (pruning), and decreasing their bit-width (quantization).
As a result, the compressed model is faster and easier to adopt in highly constrained hardware environments. 
Experiments performed on popular multivariate anomaly detection benchmarks, show that our method is capable of achieving significant model compression ratio (between 80\% and 95\%) without a significant reduction in the anomaly detection performance.

\end{abstract}


\section{Introduction}
Multivariate time series anomaly detection is a very popular machine learning problem in many industry sectors. Therefore, many research works have been proposed in this field \cite{malhotra2016lstmbased,zhou,zhang2019,su2019,USAD,tariq2019,tuor2017,zheng}. Recent works highlight that the best results in terms of detection accuracy are achieved with deep autoencoders \cite{faber2021,pietron2022_n} based on convolutional layers. Other models with satisfactory results are autoencoders based on graph neural networks \cite{graph-nn} and recurrent layers \cite{nasa-lstm,garg,malhotra2016lstmbased,Chen2019SequentialVF}.
It is worth noting that their effectiveness depends heavily on the specific  characteristics of the dataset they are assessed on. Neuroevolution provides a valuable way to address this issue, with the potential to extract optimized models for any given dataset. A notable example is the AD-NEv framework \cite{pietron2023ad}, which supports multiple layer types: CNN-based, LSTM based and GNN-based.

One potential burden of deep autoencoder models is that each additional layer or channel inside the layer slows down the training and inference process, which negatively affects their efficiency in real-time or embedded systems. In fact, any delay in their inference can have a significant impact on the operation of the  reliability of these systems. 
To this end, compression algorithms can significantly help in reducing the number of CPU cycles required to process input data.
The second advantage of their adoption is the memory footprint reduction they provide. This aspect is extremely important in cases where models are exploited in dedicated hardware e.g. IoT, edge, etc.

Many works focus on compression for deep learning   \cite{han2015learning,motaz2020,pietron2020retrain,pietronCANDAR,renda2020,xu2021,gysel2016ristretto,linda2016}, particularly for image-based data and natural language processing. The most efficient techniques are pruning \cite{han2015learning,xu2021,renda2020,frankle2019} and quantization \cite{gysel2016ristretto,han2015learning,motaz2020}. The pruning process presented in these works can be divided into structured pruning and unstructured pruning. The quantisation can be linear or non-linear.
These works show that many deep learning models may present redundant weights which can be removed without any significant drop in detection accuracy. 

Additionally, given the robustness of these models to noise in input data, weights can be quantized to a lower bit format, further decreasing the memory footprint. 
However, studies focusing on reducing the complexity of deep learning models are still limited in the anomaly detection field. To the best of our knowledge, there is no work devoted to compressing models on multivariate anomaly detection benchmarks.


To this end, in this paper we propose a compression workflow based on pruning and quantization. We adopt convolutional and graph autoencoders which have shown to be the most robust model architectures in anomaly detection tasks. In our work, pruning is incorporated in the training process, while linear and non-linear quantization based on nearest neighbour rounding is run on pruned and pre-trained models.
%
%
Our experiments leveraging state-of-the-art base model architectures \cite{faber2021,pietron2022_n} show that compression techniques like pruning and quantization can significantly reduce the complexity of deep model architectures in multivariate anomaly detection tasks. 

\begin{figure*}[ht]
   \centering

  \includegraphics[scale=0.60]{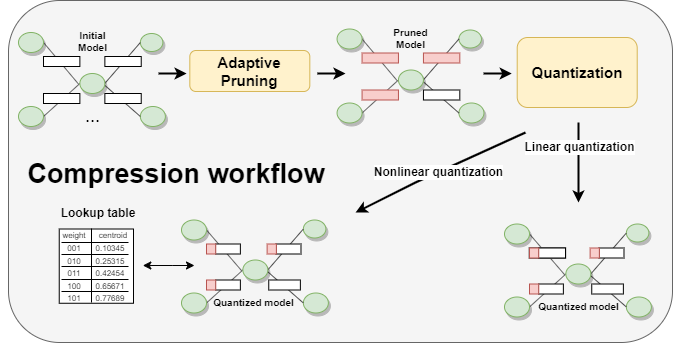}
  \caption{
  Proposed compression workflow for deep autoencoder models involving adaptive pruning and linear/nonlinear quantization stages.}
    \label{fig:algorithm}
\end{figure*}

\section{Related works}

In recent years, many efforts where directed at tackling anomaly detection problems. 
Deep learning models have shown to outperform standard machine learning models in many domains \cite{garg,faber2021,pietron2022_n}.
Examples of CNN-based approaches in the literature include temporal convolutional network AE (TCN AE) \cite{garg}, where the encoder is built from a stack of temporal convolution (TCN) \cite{bai2018empirical} residual blocks. In the decoder, convolutions in the TCN residual blocks are replaced with transposed convolutions. 
Authors in \cite{faber2021,pietron2022_n} have shown that  highly accurate CNN-based auto-encoder model architectures can be effectively optimized via neuroevolution, achieving state-of-the-art results in popular multivariate anomaly detection datasets such as SWAT, WADI, SMAP, and MSL. 

Among graph-based approaches, the authors in \cite{graph-nn} proposed a dense graph neural architecture where each node is a feature, and edges represent data exchanges between nodes. 
The work in \cite{feng2022full} introduced a full graph autoencoder for semi-supervised anomaly detection applied to large-scale IIoT systems, where conventional graph layers are replaced with variational layers model to fully capture the graph representation.
A mirror temporal graph autoencoder framework for anomaly detection is proposed in \cite{ren2024graph}, where temporal and graph convolutions with recurring units and Gaussian kernel functions are adopted to model hidden node-to-node interactions.

Considering the superior performance showcased by convolutional and graph-based autoencoder architectures in recent research works, in the present study, we adopt such models and benchmark anomaly detection datasets to assess the effectiveness of our compression approach.




Shifting the focus to model compression techniques, research on pruning techniques has shown that, in many cases, a deep learning architecture can be largely compressed without excessive losses in model performances.
%
%
Some of the most popular pruning approaches include lottery ticket search \cite{frankle2019}, pruning without retraining and local search heuristics \cite{pietron2020retrain,motaz2020}, pruning based on variational dropout \cite{molchanov}, and movement pruning  \cite{movement_pruning2020,10.1145/3007787.3001163}. These methods are based on retraining masks (static or dynamic) at each epoch, which are mostly based on gradients or absolute values of model weights. 
Mitigating the complexity of convolutional operations has also become a quite popular research problem. In \cite{pietron2020retrain}, authors show that combining retraining with pruning can significantly reduce drops in accuracy caused by removing unimportant weights. To date, pruning appears to be among the most efficient solutions for accelerating and compressing of deep learning models \cite{pietron2020retrain,frankle2019,movement_pruning2020}.
%
%
%
Another popular compression technique is quantization. Several approaches proposed for deep learning models can be categorized as linear and non-linear \cite{han2015learning,linda2016,gysel2016ristretto}, clustering-based \cite{pietronCANDAR} and hybrid \cite{motaz2020}. 
Some quantization approaches involve a retraining stage \cite{han2015learning,linda2016}, whereas some others do not \cite{motaz2020}. 
An efficient quantization technique for image processing tasks has been proposed in \cite{ibm}, 
where non-linear optimization is used to tweak quantization levels, and to obtain a reduction in loss values.



\section{Method}
{

\begin{figure*}[ht]
   \centering

  \includegraphics[scale=0.55]{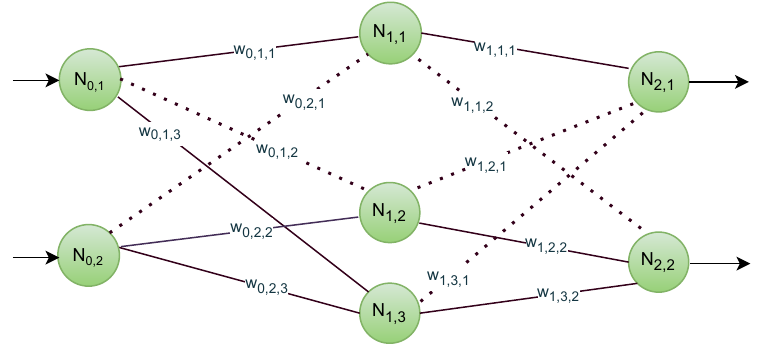}
  \includegraphics[scale=0.55]{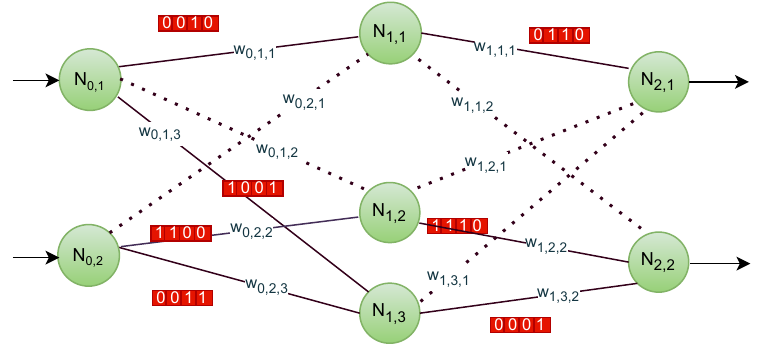}
  \caption{
  Overview of our proposed adaptive pruning and quantization approach. The initial model is pruned based on fast lottery ticket search (a), and its weights are quantized to a 4-bit representation (b).}
    \label{fig:algorithm}
\end{figure*}

In this section, we describe our proposed compression method for auto-encoder models. The rationale to focus on this particular model architecture is their robustness and effectiveness in anomaly detection tasks \cite{pietron2022_n, garg, faber2021autoencoder}.
Auto-encoders learn a compressed representation, i.e., latent space \textit{Z} of raw input data \textit{X} in an unsupervised manner. Auto-encoders are made of two parts: the encoder $E$, which transforms (encodes) the original input to the latent space, and the decoder $D$, which transforms (decodes) the latent space $Z$ to the original feature space. 
The auto-encoder architecture is a symmetrical sequence of encoder and decoder layers:
\begin{equation}
    Z = E_{\Theta}(X) = {e_{\theta_L}(e_{\theta_{L-1}}...(e_{\theta_{0}}(X)))}
\end{equation}

\begin{equation}
    D_{\Theta}(Z) = {d_{\theta'_0}(d_{\theta'_{1}}...(d_{\theta'_{L}}(Z)))}
\end{equation}


The objective of the training is the minimization of the \textit{reconstruction loss}, which corresponds to the difference between the decoder's output and the original input data. It can be expressed as:
\begin{equation}
    \mathcal{L}(X, \hat{X}) = ||X - AE_{\Theta}(X)||_2,
    \label{eq:autoencoder}
\end{equation}
where:
\begin{equation}
    AE_{\Theta}(X) = D_{\Theta}(Z),   Z = E_{\Theta}(X).
\end{equation}

The full model can also be defined in a unfolded layer-wise manner as follows:
\begin{equation}
AE_{\Theta}(X) = D_{\Theta}(E_{\Theta}(X)) =  {d_{\theta'_0}(d_{\theta'_{1}}...(d_{\theta'_{L}}(e_{\theta_{L}}(e_{\theta_{L-1}}...e_{\theta_0}(X))))},
\label{eq:model}
\end{equation}



Our compression method consists of three stages: pruning, quantization, and non-gradient fine-tuning.
The pruning stage aims to limit the number of weights in the model, identifying the most important weights for the anomaly detection task. 
Quantization reduces the bit width of all weights in the model, resulting in a further reduction of the model's size. The symbiotic combination of pruning and quantization can ensure a better effectiveness of compression.
As the last step, we leverage the non-gradient fine-tuning of the quantized model to improve the final performance.
We describe each step in detail in the following subsections.


\subsection{Pruning}
The goal of the first stage is to carry out a pruning process, which allows the retention of the most relevant parameters, thus saving computational resources involved for model inference.
The general idea is to identify a subset of weights that yield a similar anomaly detection performance to the full model. By doing so, it is possible to discard the remaining weights and reduce the model size, which positively impacts the efficiency of the model and facilitates its provisioning in resource-limited environments such as edge and IoT.
To achieve this goal, in this stage we devise pruning algorithms that involve: \textit{i)} identification of a separate sparsity level for each layer, and \textit{ii)} pruning with model retraining, to foster a more effective identification of the sparsity levels \cite{pietron2020retrain}.
%



The representation of the pruned model $AE_{\Theta}^{p}$ 
is a tuple $AE_{\Theta}^{p} = (AE_{\Theta}, M)$,  where $AE_{\Theta}$ is the original model composed by convolutional, fully-connected or LSTM layers $e_{\theta_{i}}$ and $d_{\theta_{i}}$, arranged in the specified order.
It is worth noting that our framework supports different types of layers, such as convolutional, fully-connected, and LSTM layers, among others. 
The weights for each layer are represented by the $\Theta$ tensor consisting of parameters from the encoder and decoder:
\begin{equation}
\Theta = \{\theta_{0}, \theta_{1},...,\theta_{L}, \theta'_{L},...,\theta'_{1}, \theta'_{0}\}
\label{eq:theta_}
\end{equation}

The mask tensor $M$ contains $'0'$ and $'1'$ entries, which denote, for a given layer, weights that are either pruned or retained, respectively. The tensor can be defined as a set of layer-wise masks:
\begin{equation}
M = \{M_{e_{\theta_0}}, M_{e_{\theta_1}}, \dots, M_{e_{\theta_L}}, M_{d_{\theta_L}}, \dots, M_{d_{\theta_1}}, M_{d_{\theta_0}}\}.
\label{eq:mask_}
\end{equation}
%
%
Each mask $M_{i}$ is assigned to a specific  layer, where each entry in $M_{i, j}$ can be regarded as:
\begin{equation}
M_{i, j} =
  \begin{cases}
    0  &   \quad \text{if weight is pruned}\\
    1  & \quad \text{if weight is not pruned}.
  \end{cases}
\label{eq:mask_def}
\end{equation}

We note that, in our work, a \textit{sparsity level} $\upsilon_i$ for a layer $i$ is defined as the ratio between the number of utilized weights and the total number of weights at that layer:
\begin{equation}
\upsilon_i = \frac{\sum_{j} M_{i, j}}{|M_{i}|}
\label{eq:mask___}
\end{equation}

%
We can find a threshold $\epsilon_i$ which ensures to retain the proper of weights (having a value greater than $x$) according to the sparsity $\upsilon_i$.
\begin{equation}
\label{eq:epsilon}
    \epsilon_i = x \ \ , \ \ \text{where} \ \ \frac{|abs(\theta_i) > x|}{|\theta_i|} = \upsilon_i.
\end{equation}

We leverage $\epsilon_i$ to identify the strongest weights which should be retained for a given layer:
\begin{equation}
\label{eq:mask}
M_{i, j} =
  \begin{cases}
    0  &   \quad \text{if  } abs(\theta_{i, j}) < \epsilon_{i}\\
    1  & \quad \text{if  } abs(\theta_{i, j}) > \epsilon_{i}
  \end{cases}
\end{equation}









The weighted sparsity is computed as a sum of two ratios which define the local sparsity for the encoder and decoder counterparts of the model, respectively: 
%
%
%
%
\begin{equation}
sparsity_{ws} = \sum_{j=0}^{L} \frac{ |\theta_j| \cdot \upsilon_j}{|\Theta|} + \sum_{j=0}^{L} \frac{ |\theta'_j| \cdot \upsilon_j}{|\Theta|}.
\label{eq:w_sparsity}
\end{equation}
}


The input parameters of the Algorithm \ref{alg:pruning} are: 
$AE$ - the pretrained auto-encoder,
$O$ - type of the optimizer and its parameters,
$B$ - number of batches,
$P_{S}$ - population size of the pruning candidates,
$N$ - number of epochs in final stage of pruning,
$N_{it_e}$ - number of epochs in lottery ticket search, $V_{MIN}$ = $[min_0, min_1,...,min_L, min'_L,...,min'_1,min'_0]$ - is a vector with minimum sparsity levels in all layers in a pruning mask, $V_{MAX}$ = $[max_0, max_1,...,max_L, max'_L,...,max'_1,max'_0]$ - is a vector with maximum sparsity levels in all layers in a pruning mask.


\begin{algorithm}
\begin{algorithmic}[1]

\REQUIRE{$AE$, $P_{S}$, $V_{MIN}$, $V_{MAX}$, $N$, $N_{it}$, $O$, $B$}
\REQUIRE{$P_{S}$ }
\STATE{$\Lambda$ $\gets$ $P_S$ copies of $AE$}
\STATE{$K$ $\gets$ $\emptyset$ \ \ \ \ \ \ // List of F1-Scores for all models in population $\Lambda$}
\FOR{$i = 0$ \TO $P_{S}$}
\STATE{$M$ $\gets$ $\emptyset$}
\FOR{$l = 0$ \TO $2 \cdot L$}
\STATE{$min_l$ $\gets$ $V_{MIN}[l]$}
\STATE{$max_l$ $\gets$ $V_{MAX}[l]$}
\STATE{$mean_l$ $\gets$ $min_l$ + $\frac{max_l-min_l}{2}$}
\STATE{$std_l$ $\gets$ $mean_l$ + $\frac{max_l-min_l}{6}$}
\STATE{$\upsilon_l$ $\gets$ bound(sample($\mathcal{N}(mean_l, std_l)$), $min_l$, $max_l$) \ \ \ \ \ \ // Force sparsity level in the range $(min_l, max_l)$ }
\STATE{$\epsilon_l$ $\gets$ compute threshold based on $\upsilon_l$ (see Equation \ref{eq:epsilon})}
\STATE{$M_l$ $\gets$ generate mask based on $\epsilon_l$ (see Equation \ref{eq:mask})}
\STATE{$M$ $\gets$ $M$ $\cup$ $M_l$} 
\ENDFOR
\FOR{$epoch = 0$ \TO $N_{it}$}
\FOR{$b=0$ \TO $B$}
\STATE{$\Theta$ $\gets$ train batch $b$ with $O$}
\FOR{$l = 0$ \TO $2 \cdot L$}
\STATE{$\theta_l$ = $\theta_l$ $\odot$ $M_l$ \ \ \ \ \ \ // Prune weights for layer $l$ according to mask}
\ENDFOR
\ENDFOR
\ENDFOR
\STATE{$\Lambda_i$ $\gets$ ($AE_{\Theta}$, $M$, $O$)}
\STATE{$F_1$ $\gets$ evaluate $\Lambda_i$}
\STATE{$K$ $\gets$ $K$ $\cup$ $F_1$}
\ENDFOR
\STATE{$i$ $\gets$ argmax($K$)}
\STATE{($AE_{\Theta}$, $M$, $O$) $\gets$ $\Lambda_{i}$}
\STATE{}
\STATE{// Optimize best performing model}
\FOR{$epoch = 0$ \TO $N$} 
\FOR{$b=0$ \TO $B$}
\STATE{$\Theta$ $\gets$ train batch $b$ with $O$}
\FOR{$l = 0$ \TO $2 \cdot L$}
\STATE{$\theta_l$ = $\theta_l$ $\odot$ $M_l$ \ \ \ \ \ \ // Prune weights for layer $l$ according to mask}
\ENDFOR
\ENDFOR
\ENDFOR

\end{algorithmic}
\caption{Pruning - main algorithm}
\label{alg:pruning}
\end{algorithm}

Algorithm \ref{alg:pruning} starts with the initialization of the model population (line 1), and it sets up the empty list for pruned model quality metrics (line 2). Then, in a loop, the models are pruned using a retraining process (lines 3--27). 
At the beginning of the loop, the algorithm goes through the auto-encoder layers (see internal loop: lines 5--15) and sets up the sparsity levels for each encoder and decoder layer. 
The sparsity levels are generated by normal distribution using specified mean and standard deviation (line 11). These parameters are computed based on predefined sparsity boundaries given as input parameters (line 9 and 10). When the sparsity is computed the algorithm uses eq.\ref{eq:mask_def} to define $\epsilon_l$ value for each layer (line 12). After that the layer mask is set (line 13). The next internal loop is responsible for short training with predefined $N_{it}$ epochs (lines from 16 to 23).  
The batch training is performed (lines from 17 to 22). After each batch, the chosen layer weights are set to zero using the element-wise multiplication with the layer mask (eq. \ref{eq:w_mask}, line 20):

\begin{equation}
\begin{split}
   \theta_{i} = \theta_{i} \odot M_{i}
   \end{split}
   \label{eq:w_mask}
\end{equation}

`The next lines 24 and 25 run the evaluation of the shortly pruned model. The achieved F1 metric is added to the list (line 26). At the end in lines from 30 to 37, the model with the best accuracy from the population is taken to the final long-term training with $N$ epochs (model is taken with its mask tensor, $N$>>$N_{it}$).  

\subsection{Quantization}

The second step of our methodology is the quantization process, which aims to reduce model complexity. unlike pruning, quantization does not include additional retraining. It reduces values from a floating point to a format having fewer bits. In this paper, we present two types of quantization linear and non-linear.

The linear quantization can be seen as a mapping from a floating-point data $x\in {\mathcal S}$ to a fixed-point $q\in\mathcal Q$ using a function $f_{\mathcal Q}: \mathcal S\rightarrow \mathcal Q$ as follows (assuming signed representation):

%
\begin{equation}
	\label{eq:quant}
	q = f_{\mathcal Q}(x) = \mu + \sigma \cdot \text{round}(\sigma^{-1}\cdot (x-\mu)). 
\end{equation}
where $\mu=0$ and $\sigma = 2^{-{\bf frac\_bits}}$, where fractional bit width is defined as: 
\begin{equation}{\bf frac\_bits}={\bf total\_bits}-{\bf int\_bits}-1.
\end{equation}

and integer bit-width as:

\begin{equation}
	{\bf int\_bits}=\text{ceil}(\log_{2}(\max_{x\in{\mathcal S}} |x|)) 
\end{equation}
The scaling factor $\sigma$ is essentially a shift up or down. 

The second type of quantization that we present in this paper is the non-linear method inspired by \cite{pietron2023_n}. We present the full procedure in Algorithm \ref{alg:quant}. At first, we cluster the weights to a specified number of clusters $\omega$ leveraging the KMeans procedure (line 4). Then, we assign an identifier of the cluster to each weight in a layer, choosing the cluster closest to the original value (lines 5-7). The next step involves quantization of the centroids of the clusters to $\psi$ bit-width format (lines 8-10). Then, we create a codebook, which contains a mapping between each original weight and quantised cluster's centroid $w_q$ (line 12).

The similar approach can be found in \cite{pietronCANDAR, pietron2023_n} when the method is used for fast compression of the NLP models and for GPU-based DL models acceleration, respectively.

\begin{algorithm}
\begin{algorithmic}[1]
\REQUIRE{$\omega$, $\psi$}
\STATE{$L$ $\gets$ get model fc and conv layers}
\STATE{$c_q$ $\gets$ empty list} //initial list for quantised centroids
\FOR{$l$ $\textbf{in}$ $L$}
\STATE{$C$, $\theta_{ca}$ $\gets$ KMeans($\theta_l$, $\omega$) //$C$ is a list of centroids, $\theta_{ca}$ is a list of weights to centroids assignment}
\FOR{$w$ $\textbf{in}$ $l$}
\STATE{$w$ $\gets$ $\theta_{ca}$[$w$]}
\ENDFOR
\FOR{$c$ $\textbf{in}$ $C$}
\STATE{$c_q$[c] $\gets$
quantise\_centroids($c$, $\psi$) //eq.20-22}
\ENDFOR
\FOR{$w$ $\textbf{in}$ $l$}
\STATE{$w_q$ $\gets$ $c_q$[$w$]}
\ENDFOR
\ENDFOR
\caption{Nonlinear quantization, \cite{pietron2023_n}}
\label{alg:quant}
\end{algorithmic}
\end{algorithm}

The complexity of the auto-encoder can be  defined as the size of the hardware blocks required to implement the MAC (multiply-accumulation) operations. The capacity of the model is the amount of memory needed for weights storage. 
The number of MAC operations in the layer is equal to $P_i$ $\cdot$ $C_i$ $\cdot$ $D_i$ $\cdot$ $H_i$ $\cdot$ $W_i$, where $P_i$ is the number of neurons in an output feature map. The memory footprint (capacity) of the single layer is $C_i$ $\cdot$ $D_i$ $\cdot$ $H_i$ $\cdot$ $W_i$. The pruning reduces MAC operations and capacity by factor which is equal sparsity level ($sparsity_i$). In case of quantization the capacity is decreased linearly. The MAC operations in 8-bit format are reduced from floating point counterpart by 1/9. 
Reducing bit-width of the weights further increases this ration linearly.

\section{Results}

The research questions posed by our paper are the following:

\textbf{RQ1.} How efficient dynamic pruning can be in anomaly detection auto-encoder architectures? 

\textbf{RQ2.} How effectively can deep state-of-the-art anomaly detection models be reduced by means of quantization?

\textbf{RQ3.} What is the efficiency of linear and non-linear quantization on a pretrained autoencoder?

{
In our experiments, we consider state-of-the-art architectures in recent benchmarks for anomaly detection \cite{pietron2023ad}\cite{faber2021ensemble}, i.e. convolutional autoencoders (CNN AE) and graph-based autoencoders (GDN).
We adopt popular benchmark datasets: SWAT, WADI-2019, MSL, SMAP. The CNN AE for
SWAT and WADI-2019 consist of 6 layers, for SMAP and WADI they have 12 layers.
The implementation of our pruning and quantization approach is done in Python and PyTorch. All our experiments were executed on a workstation equipped with  Nvidia Tesla V100-SXM2-32GB GPUs. The $V_{MIN}$ and $V_{MAX}$ parameters were set to 0.2 and 0.8, respectively. The population size $P_{S}$ was set to 16. In the pruning experiments the architectures of these models were taken and trained from scratch by Algorithm \ref{alg:pruning}. The linear and nonlinear quantization were run on pretrained models. In all quantization experiments the output neuron activations are in 8-bit format.

Results in Table \ref{tbl:baseline-models} show the performance achieved with baseline models (without pruning and quantization) and all the considered datasets. The results achieved by CNN AE are the best among all models tested on the analyzed datasets. The GDN achieves the second result in the case of the WADI-2019 and SWAT.


Results in Table \ref{tbl:pruning-sparsity} show the performance of models following the pruning stage of our proposed compression workflow with different Sparsity levels. We observe that with a sparsity level of $0.2$ the anomaly detection performance drops slightly in SWAT: from 82.0 to 81.45 (CNN AE) and from 81.0 to 80.51 (GDN). The performance drop is more significant for WADI-2019: from 62.0 to 56.28 (CNN AE) and from 57.0 to 53.5 (GDN). These results show that is difficult to reduce significant number of weights for both models on WADI-2019. The drop for Sparity level $0.2$ can be acceptable for SWAT (~$0.5$). The higher Sparsity level increases the drop further for both datasets. 
In case of SMAP and MSL when Sparsity increases to $0.75$, the performance is still at the same level as in baseline models (for both CNN AE and GDN). 
These surprising results can be motivated as pruning can, in some cases, provide a noise reduction capability in the presence of noisy data in multivariate datasets, resulting in a more robust model. 
Overall, our experimental result show that pruning can be an effective strategy to compress deep autoencoder models for anomaly detection, especially for MSL and SMAP datasets (\textbf{RQ1}). 

Results in Table \ref{tbl:linear-quantization} show the performance of models following the quantization stage of our proposed compression workflow with different bit width configurations (16-bit, 8-bit, 5-bit and 4-bit).
Overall, our experimental results show that 16-bit and 8-bit quantization can be quite effective in reducing the complexity of deep autoencoder models used for anomaly detection tasks (\textbf{RQ2}). In case of 5-bit and 4-bit quantization there is significant drop on WADI-2019 and SWAT. Both models CNN AE and GDN are robust for 4-bit quantization and give the F1-score at the same level as the baseline counterparts.

Results in Table \ref{tbl:nonlinear-quantization} show the performance of models which were quantized with non-linear quantization. It can be observed that for nonlinear 16-bit and 8-bit there is no drop in accuracy for both models and datasets (\textbf{RQ2}). The drop in case of 4-bit quantization is acceptable only for MSL and SMAP.

The research results presented in \cite{pietron2023_n} show that sparse 1D convolutional layers can be speed up on GPU using sparse convolution. The sparsity above 70\% guarantees that sparse convolution outperforms standard CuDnn implementation. Additionally, it shows that GPU can make usage from reduced precision format. These two aspects allows to improve models time efficiency on GPU (\textbf{RQ3}). The GDN model consists of convolution, linear and attention operations. In this case, reducing the representation of weights can also be used to increase the speed of their operation on the GPU without loss of accuracy. 


}

\begin{table}[H]
\setlength{\tabcolsep}{12pt}
\centering
\caption{Baseline results (no pruning, no quantization) in terms of F1-Score for all models and datasets.}
\begin{tabular}{|c|c|c|c|c|}
\hline
\textbf{Datasets} & \textbf{CNN AE} & \textbf{GDN} \\ \hline
\textbf{SWAT}     &  82.0     &   81.0                                     \\ \hline
\textbf{WADI-2019}     & 62.0  &  57.0                                   \\ \hline
\textbf{MSL}     &  57.0       &  30.0                                       \\ \hline
\textbf{SMAP}     &  77.0       &  33.0              \\ \hline
\end{tabular}
\label{tbl:baseline-models}
\end{table}







\begin{table}[H]
\setlength{\tabcolsep}{10pt}
\centering
\caption{Model performance in terms of F1-Score with proposed pruning workflow and different sparsity levels applied to each layer.}
\begin{tabular}{|c|c|c|c|c|}
\hline
& \multicolumn{2}{c}{Sparsity=0.2} & \multicolumn{2}{c|}{Sparsity=0.75} \\ 
\hline
\textbf{Datasets} & \textbf{CNN AE} & \textbf{GDN}  & \textbf{CNN AE} & \textbf{GDN}\\ \hline
\textbf{SWAT}     &  81.45   & 80.51  & - & - \\ \hline
\textbf{WADI-2019}     & 56.28    &  53.5  & - & - \\ \hline
\textbf{MSL}        &    -    &   -  & 57.01    & 30.2     \\ \hline
\textbf{SMAP}       &   -    &  -  & 77.02    &  32.9      \\ \hline
\end{tabular}
\label{tbl:pruning-sparsity}
\end{table}


\begin{table}[H]
\setlength{\tabcolsep}{9pt}
\centering
\caption{Experimental results (F1-score) with linear quantization and different bit-width configurations.}
\begin{tabular}{|c|c|c|c|c|}
\hline
& \multicolumn{2}{c}{16-bit} & \multicolumn{2}{c|}{8-bit} \\ 
\hline
\textbf{Datasets} & \textbf{CNN AE}   & \textbf{GDN}  & \textbf{CNN AE}   & \textbf{GDN} \\ \hline
\textbf{SWAT}     &     81.90 
& 80.80 & 81.98 
& 80.70 \\ \hline
\textbf{WADI-2019}     &   62.13    & 56.90    &       62.06 
&   56.80                                                 \\ \hline
\textbf{SMAP}       &     77.15   & 32.85 &     77.05     & 32.82  \\ \hline
\textbf{MSL}        &    57.21   & 29.95   &    57.14      & 29.91   \\ \hline
& \multicolumn{2}{c}{5-bit} & \multicolumn{2}{c|}{4-bit} \\ 
\hline
\textbf{Datasets} & \textbf{CNN AE}   & \textbf{GDN}  & \textbf{CNN AE}   & \textbf{GDN}\\ \hline
\textbf{SWAT}     &     80.70 
&  79.80 &  16.44                         &         78.20                \\ \hline
\textbf{WADI-2019}  &     54.52    & 54.5 &    48.20                   &       45.5            \\ \hline
\textbf{SMAP}       &      76.1    &  32.65  &   75.6  & 32.5    \\ \hline
\textbf{MSL}        &    56.45    &  29.69    &    56.1   & 29.55  \\ \hline
\end{tabular}
\label{tbl:linear-quantization}
\end{table}

\begin{table}[H]
\setlength{\tabcolsep}{9pt}
\centering
\caption{Experimental results (F1-score) with non-linear quantization and different bit-width configurations.}
\begin{tabular}{|c|c|c|c|c|}
\hline
& \multicolumn{2}{c}{16-bit} & \multicolumn{2}{c|}{8-bit} \\ 
\hline
\textbf{Datasets} & \textbf{CNN AE}   & \textbf{GDN}  & \textbf{CNN AE}   & \textbf{GDN} \\ \hline
\textbf{SWAT}     &     80.36                         &      80.70       &     80.27   &      80.75                                             \\ \hline
\textbf{WADI-2019}     &    57.56                     &   55.0  &      59.11                   &          55.5                  \\ \hline
\textbf{SMAP}       &   77.0   & 32.95  &   76.8  &   32.92   \\ \hline
\textbf{MSL}        &    56.97    & 29.91   &      56.98   &   29.92  \\ \hline
& \multicolumn{2}{c}{5-bit} & \multicolumn{2}{c|}{4-bit} \\ 
\hline
\textbf{Datasets} & \textbf{CNN AE}   & \textbf{GDN}  & \textbf{CNN AE}   & \textbf{GDN}\\ \hline
\textbf{SWAT}     &     78.45                          &    33.5    &     16.35                         &         25.0       \\ \hline
\textbf{WADI-2019}     &       17.87                  &   31.0   &      10.89                  &        25.0              \\ \hline
\textbf{SMAP}       &    76.2  & 32.71   &  75.9    &  32.63   \\ \hline
\textbf{MSL}        &    56.1    & 29.67   &     56.90    &  29.63  \\ 
\hline
\end{tabular}
\label{tbl:nonlinear-quantization}
\end{table}

\section{Conclusions and future works}

In this paper we proposed a compression workflow leveraging  pruning and quantization stages.
While pruning is incorporated in the training process, linear and non-linear quantization is performed on pruned and pre-trained models.
Our experiments leveraging state-of-the-art convolutional and graph autoencoder model  architectures revealed the trade-off between model compression and anomaly detection performance that pruning and quantization techniques can achieve in benchmark multivariate anomaly detection settings. 
Among key findings, we observed that pruning can be quite effective with MSL and SMAP datasets, and 16-bit and 8-bit quantization only impacted in a small drop in terms of F1 score. Additionally, the 4-bit quantization gives the same accuracy levels as in floating point mode. On the other hand, pruning was not effective with the WADI dataset. The presented results show that anomaly detection autoencoders can be reduced from 80\% (8-bit quantization and 20\% sparsity level, WADI-2019 and SWAT) to about 95\% (4-bit quantization and 75\% sparsity level, MSL and SMAP).  

Future work will focus on more advanced quantization techniques based on model retraining, which could  decrease the drop in F1-Score for very low bit-widths. 


\bibliographystyle{unsrt}  
\bibliography{references}

\end{document}